\documentclass{ifacconf}

\usepackage{graphicx}      
\usepackage{natbib}        

\usepackage{amsmath}
\usepackage{amsfonts}       

\theoremstyle{definition}
\newtheorem{definition}{Definition}

\theoremstyle{assumption}

\theoremstyle{pro}
\newtheorem{pro}{Proposition}

\usepackage{comment}
\usepackage{xcolor}

\newcommand{\R}{\mathbb{R}}
\newcommand{\E}{\mathbb{E}}
\newcommand{\Z}{\mathbb{Z}}

\newcommand\newtag[2]{#1\def\@currentlabel{#1}\label{#2}}

\begin{document}
	\begin{frontmatter}
		
		\title{Estimating  Koopman operators for nonlinear dynamical systems: a nonparametric  approach} \thanks[footnoteinfo]{Support by the Italian Ministery for Higher Education (MIUR) under the program PRIN 2017 is gratefully acknoweledged.}

\author[First]{Francesco Zanini} 
\author[Second]{Alessandro Chiuso} 

\address[First]{Department of Information Engineering, University of Padova, Padova 35131, Italy (e-mail: francesco.zanini.3@phd.unipd.it).}
\address[Second]{Department of Information Engineering, University of Padova, Padova 35131, Italy (e-mail: chiuso@dei.unipd.it)}
		
		\begin{abstract}                
			The Koopman operator is a mathematical tool that allows for a linear description of non-linear systems, but working in infinite dimensional spaces. Dynamic Mode Decomposition and Extended Dynamic Mode Decomposition are amongst the most popular finite dimensional approximation. In this paper we capture their core essence as a dual version of the same framework, incorporating them into the Kernel framework. To do so, we leverage the RKHS as a suitable space for learning the Koopman dynamics, thanks to its intrinsic finite-dimensional nature, shaped by data. We finally establish a strong link between kernel methods and Koopman operators, leading to the estimation of the latter through Kernel functions. We provide also simulations for comparison with standard procedures.
					\end{abstract}
		
		
	\end{frontmatter}
	
	\section{Introduction}
	The Koopman operator represents an alternative description for dynamical systems, first introduced in the seminal work by \cite{koopman1931hamiltonian}. The interest in this perspective has recently been renewed thanks to the fundamental work by \cite{mezic2005spectral}, which deals with the problem of decomposing the evolution of a vector field from the perspective of operator theory, and to the key paper \cite{rowley2009spectral}, introducing Koopman modes, which represent the collective motion of fluids. This line of work has always had a strong connection with the world of fluid dynamics, being one of the most straightforward field of application and motivation, however in more recent years a link with Kernel methods for system identification started to emerge from the literature. \\
	In particular, works such as \cite{kevrekidis2016kernel}, \cite{kawahara2016dynamic}, and \cite{das2020koopman} address the problem of estimating Koopman modes in a larger space, thus naturally introducing the RKHS to generalise the standard Krylov subspace. Nonetheless so far Kernel methods have just been used as a convenient description of an infinite dimensional space of functions, and no theoretical bridge has yet been established between the two frameworks. Also, in the system identification literature, the Koopman operator has been used as a tool to estimate non-linear dynamical models. For instance in \cite{mauroy2019koopman} it has been used to estimate continuous time dynamical systems, exploiting the linear representation provided by the Koopman lifting. \\
	In this work we introduce a nonparametric framework to estimate Koopman operators in infinite dimensional spaces. We do so by exploiting the language of Reproducing Kernel Hilbert Spaces (RKHS) and Gaussian Processes (GP). Our contribution also include a comparative discussion with standard approaches, namely Dynamic Mode Decomposition (DMD) and Extended Dynamic Mode Decomposition (EDMD). Simulation results are included, showing both the reconstruction of a non-linear state evolution map as well as the update of a simple observable. \\
	The structure of the paper is as follows: Section \ref{prel} provide a brief background on RKHS and Koopman operators, including DMD and EDMD.  Section \ref{dual} discusses the latter two approaches in their context of so-called values-based and function-based Koopman reconstruction. Section \ref{RKHS} represents the core of our paper and introduces the infinite dimensional framework. Section \ref{simulations} contains some numerical experiments and Conclusions are drawn in Section \ref{conclusions}.
	
	\section{Preliminaries}\label{prel} 
	
		\subsection{Kernel methods}
		Consider the general problem of estimating an unknown function $\boldsymbol{f}(\cdot)$ given by:
	\begin{equation*}
		\boldsymbol{y} = \boldsymbol{f} \left( \boldsymbol{x} \right) 
	\end{equation*}
	from given noisy input-output pairs $\left\lbrace \boldsymbol{x}_i, \boldsymbol{y}_i\right\rbrace_{i=1}^M$. \\
	Even by neglecting the noise, this problem is ill-posed as there are infinite functions which could explain the data: in order to obtain a well-defined solution, we restrict the search space to an RKHS, which is an Hilbert space that can be defined by selecting a particular Kernel function.
	
	The problem can then be recast into:
	\begin{equation*}
		\mathop{\rm min}_{\boldsymbol{f} \in \mathcal{H}} \left\lbrace \frac{1}{M} \sum_{i=1}^M L \left( \boldsymbol{y}_i, \boldsymbol{f} \left( \boldsymbol{x}_i \right) \right) + \lambda \Vert \boldsymbol{f} \Vert_{\mathcal{H}} \right\rbrace 
	\end{equation*}
	for any convex loss function $L(\cdot, \cdot)$ and regularization parameter $\lambda$. The term $\Vert \boldsymbol{f} \Vert_{\mathcal{H}}$ represents the norm of the function in the RKHS. \\
	By the Representer theorem, the solution of this problem is known to lie in a finite dimensional subspace, spanned by the Kernel sections, i.e.:
	\begin{equation*}
		\boldsymbol{f} \left( \boldsymbol{x} \right) = \sum_{i=1}^{M} \alpha_i K(\boldsymbol{x}, \boldsymbol{x}_i)
	\end{equation*}
	which is the best reconstruction according to the chosen kernel. The kernel can also be interpreted from a Bayesian perspective; in fact, the same solution would be obtained by assuming that $\boldsymbol{f}(\cdot)$ is a Gaussian Process (\cite{Rasmussen}) with zero mean and covariance function $K(\cdot,\cdot)$. 
	\subsection{Basics of Koopman operator}
	Consider an autonomous discrete-time dynamical system in state-space form:
	\begin{equation} \label{eq:auto_system}
		\boldsymbol{x}_{t+1} = \boldsymbol{f} \left( \boldsymbol{x}_{t} \right), \quad t \in \Z^+, \quad \boldsymbol{x}_{t} = \boldsymbol{x}_0,
	\end{equation}
	the state transition function $\boldsymbol{f}$ maps the state space $ \mathbb{R}^{n}$ in itself. 

	Given a class of scalar, complex  
	valued functions $ \mathcal{F}$, we also define the notion of \emph{observables} as follows:  	\begin{definition}
		An \textit{observable} is any function $\psi(\cdot) : \mathbb{R}^{n} \rightarrow \mathbb{C}, \ \psi \in \mathcal{F}$ mapping the state of the system into a scalar. 
	\end{definition}
	
	The value taken by the observable is uniquely determined by the  state of the system and its evolution over time 
 is described by the composition of the observable with the state dynamics $\boldsymbol{f}(\cdot)$, i.e.:
 \begin{equation}
 		\psi \left( \boldsymbol{x}_{t+1} \right) = \psi \left( \boldsymbol{f} \left( \boldsymbol{x}_{t} \right) \right) \label{eq:obs_map}
	\end{equation}
	
	It is useful to observe that, provided a sufficiently rich set of observables $\psi_i(\boldsymbol{x})$, $i=1,..,N$ is considered, their evolution completely characterize the state dynamics $\boldsymbol{f}(\cdot)$ (in fact, it is sufficient to take $\psi_i (\boldsymbol{x}) = x_i$, for $i=1, \dots, n$). 

	The Koopman operator is the mathematical object that describes the evolution of observables \eqref{eq:obs_map} under the state evolution, namely:
	
	\begin{definition}
		The \textit{Koopman operator} $\mathcal{U} : \mathcal{F} \rightarrow \mathcal{F}$  is defined as the mapping of a generic observable into the composition of the same observable with the state transition function:
		\begin{equation*}
			\mathcal{U} \left[\psi \right](\boldsymbol{x}) = \psi \left( \boldsymbol{f} \left( \boldsymbol{x} \right) \right), \quad \forall \psi \in \mathcal{F}
		\end{equation*}
	\end{definition}
	It is worth noticing that the Koopman operator  $\mathcal{U}$ is  linear and infinite dimensional.

	We shall now describe two well-known strategies to learn Koopman operators from data.
	
	\subsection{Learning Koopman operator from data} \label{ssec:learn_koop}
	
	The problem of estimating an infinite dimensional operator from finite data is an ill-posed inverse problem. One possible solution to ill-conditioning is to seek for a finite dimensional approximation. This is the route followed by the so called \emph{Dynamic Mode Decomposition} (DMD hereafter) and   \emph{Extended Dynamic Mode Decomposition} (EDMD hereafter) that will be described in the next Sections. 
	
	Both approaches rely on a finite dimensional approximation of the observable space ${\cal F}$ induced by the choice of a finite number of basis functions $\mathcal{D}: = \{\psi_1, \psi_2, \dots, \psi_N\} $. In particular we define 
	
	$$
	\begin{array}{rcl}
	{\cal F}_\mathcal{D}&: =& {\rm span}\{{\cal D}\} =  {\rm span}\{\psi_1, \psi_2, \dots, \psi_N\}  \\
	&=& \{ \psi(\boldsymbol{x}): \psi(\boldsymbol{x})=\sum_{k=1}^N \psi_k(\boldsymbol{x}) \alpha_k; \ \alpha_k \in \R\} 
	\end{array}
	$$
	
	For future use let us also denote with  $\{(\boldsymbol{x}^{(m)}, \boldsymbol{y}^{(m)})\}_{m=1}^M$ where $ \boldsymbol{y}^{(m)} = \boldsymbol{f}(\boldsymbol{x}^{(m)})$ a finite set of observations from the dynamical system \eqref{eq:auto_system}. In addition,  given an observable $\psi(\boldsymbol{x}) = \sum_{k=1}^{N}  \psi_k (\boldsymbol{x}) \alpha_k  = \Psi(\boldsymbol{x}) \boldsymbol{\alpha}$,
	let us define 
	the matrices:
\begin{align*}
	P_x = \begin{bmatrix}
	\Psi \left( \boldsymbol{x}^{(1)} \right) \\
	\vdots \\
	\Psi \left( \boldsymbol{x}^{(M)} \right)
	\end{bmatrix} \quad P_y = \begin{bmatrix}
	\Psi \left( \boldsymbol{y}^{(1)} \right) \\
	\vdots \\
	\Psi \left( \boldsymbol{y}^{(M)} \right)
	\end{bmatrix} 	\end{align*}       
	and 
	\begin{align*}
	\psi( \bar{\boldsymbol{x}})  = \begin{bmatrix}
	\psi \left( \boldsymbol{x}^{(1)} \right) \\
	\vdots \\
	\psi \left( \boldsymbol{x}^{(M)} \right)
	\end{bmatrix}  \quad 
	  \psi( \bar{\boldsymbol{y}})  = \begin{bmatrix}
	\psi \left( \boldsymbol{y}^{(1)} \right) \\
	\vdots \\
	\psi \left( \boldsymbol{y}^{(M)} \right)
	\end{bmatrix}
	\end{align*}       
	
Let us also observe that since  $\psi(\boldsymbol{x})  = \Psi(\boldsymbol{x}) \boldsymbol{\alpha}$ we also have $ \psi( \bar{\boldsymbol{x}}) =  P_x \boldsymbol{\alpha} $ and $ \psi( \bar{\boldsymbol{y}})  =  P_y \boldsymbol{\alpha}$.

	\subsection{Dynamic Mode Decomposition}\label{sec:DMD}

	This approach simply relies on the fact that the Koopman operator is a linear operator in the lifted space of observables; by approximating observables (which are infinite dimensional objects) with their evaluations on a finite set on points, the following should hold for any observable:
	\begin{equation*}
		\psi( \bar{\boldsymbol{y}})   \simeq \hat U_{DMD} \psi( \bar{\boldsymbol{x}}) 
	\end{equation*}
	Writing this equation for all the basis functions $\psi_k(\boldsymbol{x})$, and staking columnwise,  $\hat U_{DMD}$ can be obtained by solving the least squares problem:
	\begin{equation*}
	\hat U_{DMD} = {\arg \min} \; \| P_y - U_{DMD} P_x\|^2 
	\end{equation*}
	whose solution is given by 
	 \begin{equation}\label{DMD}
	\hat U_{DMD} = P_yP_x^\dagger
	\end{equation}
	
	\subsection{Extended Dynamic Mode Decomposition}\label{sec:EDMD}

	The finite dimensional approximation of ${\cal U}$ attempts to describe the evolution of 
	observables in ${\cal F}_N$. Unfortunately in general the space ${\cal F}_N$
	is not invariant w.r.t. the Koopman evolution, i.e. given $\psi \in {\cal F}_N$: 
	\begin{equation*}
	{\cal U}[\psi] \notin  {\cal F}_N
	\end{equation*}
	and therefore, at best, we can describe the evolution of the projection of ${\cal U}[\psi]$ onto  ${\cal F}_N$.
	In this regard, let us consider the decomposition:
	\begin{equation*}
	{\cal U}[\psi](\boldsymbol{x}) = \Pi_{{\cal F}_N}\left[ {\cal U}[\psi] \right](\boldsymbol{x}) + r(\boldsymbol{x})
	\end{equation*}
	where $\Pi_{{\cal F}_N}$ denotes orthogonal projection of the observables, i.e. the element of ${\cal F}_N$ such that the norm of
	\begin{equation*}
		r(\boldsymbol{x}): ={\cal U}[\psi](\boldsymbol{x}) -  \Pi_{{\cal F}_N}\left[ {\cal U}[\psi] \right](\boldsymbol{x})
	\end{equation*}
	is minimized. 
	
	EDMD implements this idea using a finite set of observations  $\{(\boldsymbol{x}^{(m)}, \boldsymbol{y}^{(m)})\}_{m=1}^M$ where $ \boldsymbol{y}^{(m)} = \boldsymbol{f}(\boldsymbol{x}^{(m)}) $.

Now, we seek for a  for a vector $\boldsymbol{\beta}$ such that $\Psi(\boldsymbol{x}) \boldsymbol{\beta}$ is the orthogonal projection of $\psi(\boldsymbol{y}) = {\cal U}[\psi](\boldsymbol{x})$. This can be obtained as the solution of the optimization problem:
	\begin{equation*}
\boldsymbol{\beta} = {\arg \; \min} \frac{1}{m} \sum_{i=1}^m (\psi( \boldsymbol{y}^{(m)} ) - \Psi(\boldsymbol{x}^{(m)})\boldsymbol{\beta})^2
	\end{equation*}
The solution to this problem is given by:
	\begin{equation*}
	 \boldsymbol{\beta} = P_x^\dagger \psi( \bar{\boldsymbol{y}})  =  {P_x^\dagger P_y} \boldsymbol{\alpha} 
	\end{equation*}
	Therefore the matrix \begin{equation}\label{EDMD}
	\hat { U}_{EDMD}: = P_x^\dagger P_y
	\end{equation}
	
	 is the finite dimensional representation of the Koopman operator in the coordinates induced by the basis functions $\psi_k(\boldsymbol{x})$, $k=1,..,N$, i.e. it describes how any function $\psi = \Psi \boldsymbol{\alpha}$ is mapped via the system dynamics to:
	\begin{equation*}
	{\cal U}[\psi](\boldsymbol{x}) = \psi(f(\boldsymbol{x})) \simeq \Psi(\boldsymbol{x}) \boldsymbol{\beta} = \Psi(\boldsymbol{x})  \hat { U}_{EDMD}\boldsymbol{\alpha}
	\end{equation*}
	where again the approximate equality is due to the finite dimensional approximation.

	\section{A dual view of DMD and EDMD}\label{dual}
	The two approaches presented in the previous sections can be given a dual interpretation, as operators acting on function values (DMD) or in function space (EDMD). In particular, as already seen in Section \ref{sec:EDMD}
	the EDMD approximation \eqref{EDMD} is the solution of the problem 
	\begin{equation*}
			\min_{A \in \mathbb{R}^{N\times N}} \Vert P_x A - P_y \Vert^2_F
		\end{equation*}
that describes how the coefficients $\boldsymbol{\alpha}$ of a function $\psi(\boldsymbol{x})  = \Psi(\boldsymbol{x}) \boldsymbol{\alpha}$ should be mapped so that the coefficient vector given by
$\boldsymbol{\beta}:=\hat U_{EDMD} \boldsymbol{\alpha}$ provides an approximation of the mapped function ${\cal U}[\psi]$ through the relation $\Psi(\boldsymbol{x}) \boldsymbol{\beta}$.

For this reason this is called a \emph{function-based} approximation and we
		define
		$$
		U_f: = U_{EDMD}
		$$
		where the subscript $f$ stand for \emph{function}.

	Similarly, the DMD approach finds an operator $\hat U_{DMD}$ given in \eqref{DMD}, that solves the problem:
	\begin{equation*}
			\min_{A \in \mathbb{R}^{N\times N}} \Vert A P_x - P_y \Vert^2_F
		\end{equation*}
		and it is such that, given the values $\psi( \bar{\boldsymbol{x}}) $  of any observable $\psi(\boldsymbol{x})$ on the training points $\boldsymbol{x}^{(m)}$, $m=1,..,M$, it outputs an approximation of the values the mapped observable takes, i.e.
		$$
		\psi( \bar{\boldsymbol{y}})  =\psi(f( \bar{\boldsymbol{y}})) \simeq  \hat U_{DMD}\psi( \bar{\boldsymbol{x}}) 
		$$
		Note the in this second approach one never needs the observable $\psi(x)$ but only its values on the training inputs $\boldsymbol{x}^{(m)}$, on the other hand it can only approximate the observable values on the mapped points $f(\boldsymbol{x}^{(m)})$. For this reason this is also called \emph{valued based} approach and we define
		$$
		U_v: = U_{DMD}
		$$
		where the subscript $v$ stand for \emph{value}.


	\section{Infinite Dimensional hypothesis: Koopman operators in RKHS}\label{RKHS}

	In this section we establish the link between the Koopman framework and Kernel methods in Bayesian identification. In order to account for  undermodeling we assume that the system dynamics \eqref{eq:auto_system} is also affected by some noise process $\boldsymbol{\omega}_t$, so that 
	\begin{equation} \label{eq:measures}
	\boldsymbol{x}_{t+1} = \boldsymbol{f}(\boldsymbol{x}_t) + \boldsymbol{\omega}_t
	\end{equation}
	Of course the (deterministic) Koompan operator will only model the drift term $\boldsymbol{f}(\boldsymbol{x}_t) $ but its estimation will account for the fact $\boldsymbol{x}_{t+1} $ will not exactly match $ \boldsymbol{f}(\boldsymbol{x}_t) $.

We shall now see how the finite dimensional approximation used in EDMD can be reframed with the language of Kernels. Later, in Section \ref{ssec:ker_to_koop} we will directly derive an estimate of the Koopman operator starting from an  infinite dimensional hypothesis space.

	\subsection{From EDMD to Kernels} \label{ssec:koop_to_ker}
	Following for example the function-based approach we can select a dictionary of observable $\mathcal{D}$ to understand the evolution of the characteristics of the system we are interested in. The approximated Koopman operator derived in Subsection~\ref{ssec:learn_koop} gives us the coefficients of $\psi(\boldsymbol{f})$ in the chosen basis as:
	\begin{equation*}
		\boldsymbol{\beta} ={U}_f \boldsymbol{\alpha} = P_x^\dagger P_y \boldsymbol{\alpha}
	\end{equation*}
	by mapping the coefficients $\boldsymbol{\alpha}$ of a generic function $\psi \in \mathcal{F}_\mathcal{D}$ into the coefficients $\boldsymbol{\beta}$ that allows for the best representation in $\mathcal{F}_\mathcal{D}$ of the composition $\psi \circ \boldsymbol{f}$. \\
	Accounting also for the fact that due to the noise $\boldsymbol{\omega}$ in \eqref{eq:measures} we have
	$\psi(\bar{\boldsymbol{y}})  \neq  \psi( \boldsymbol{f}(\boldsymbol{x}))$ and thus we can model this mismatch by a proper perturbation $ \boldsymbol{\epsilon}$ so that 
	$$
	\psi(\bar{\boldsymbol{y}}) = \psi( \boldsymbol{f}(\boldsymbol{x})) +  \boldsymbol{\epsilon}
	$$
 We now would  like to find  $\boldsymbol{\beta}$ such that $ \psi( \boldsymbol{f}(\boldsymbol{x})) = P_x \boldsymbol{\beta}$ and therefore solve
 $$\psi(\bar{\boldsymbol{y}}) = P_x \boldsymbol{\beta} + \boldsymbol{\epsilon}
$$
for $\boldsymbol{\beta}$. To frame this problem in a Bayesian setting  we model the uncertainty  $\boldsymbol{\epsilon}$ a zero mean Gaussian term with variance 
$ \mathbb{V} \left[ \boldsymbol{\epsilon} \right] = \sigma^2I$
 and also assume a prior for $\boldsymbol{\beta}$ of the form
 $$
 \boldsymbol{\beta} \sim{\cal N}(0, \Lambda)
 $$
 Observing also that since $\psi(\boldsymbol{x}) \in {\cal F}_\mathcal{D}$ there exists $\boldsymbol{\alpha}$ such that $\psi(\bar{\boldsymbol{y}}) = P_y\boldsymbol{\alpha}$, we should find the $MAP$ estimator of $\boldsymbol{\beta}$ in the linear measurement model
 \begin{equation*}
		\psi(\bar{\boldsymbol{y}})  = P_y \boldsymbol{\alpha} = P_x \boldsymbol{\beta} + \boldsymbol{\epsilon}, 	\end{equation*}
The solution is then given by:
\begin{align*}
		\hat{\boldsymbol{\beta}} &= \Lambda P_x^\top \left(P_x \Lambda P_x^\top + \sigma^2I_M \right)^{-1} P_y\boldsymbol{\alpha} \notag \\
		&= \left( \sigma^2 \Lambda^{-1} + P_x^\top P_x \right)^{-1} P_x^\top P_y\boldsymbol{\alpha}
	\end{align*}
which can be seen as a regularized version of $EDMD$. Thus we define 
\begin{equation} \label{eq:reg:EDMD}
		U_f^{(r)} = \left( \sigma^2 \Lambda^{-1} + P_x^\top P_x \right)^{-1} P_x^\top P_y
	\end{equation}
	and the EDMD estimator in \eqref{EDMD} is recaptured setting  $\sigma^2 = 0$.

If we now express the EDMD estimate of the function ${\cal U}[\psi](\boldsymbol{x})$ we obtain:
\begin{equation}\label{EDMDBasis}\begin{array}{rcl}
\widehat{ {\cal U}[\psi]}(\boldsymbol{x}) &=& P_x \boldsymbol{\beta} = \Psi(\boldsymbol{x}) U_f^{(r)} \boldsymbol{\alpha} \\
& =& \Psi(\boldsymbol{x}) \Lambda P_x^\top \left(P_x \Lambda P_x^\top + \sigma^2I_M \right)^{-1}  P_y \boldsymbol{\alpha} \\
& = &  \Psi(\boldsymbol{x}) \Lambda P_x^\top \left(P_x \Lambda P_x^\top + \sigma^2I_M \right)^{-1} \psi\bar{\boldsymbol{y}}
\end{array}
\end{equation}
 Defining the Kernel as inner product of basis functions as:
 $$
 K(\boldsymbol{x},\boldsymbol{y}) := \Psi(\boldsymbol{x})\Lambda \Psi(\boldsymbol{y})^\top  = \left\langle \Psi(\boldsymbol{x}),\Psi(\boldsymbol{x})\right\rangle _{\Lambda}
 $$
 Equation \eqref{EDMDBasis} can be rewritten as:
 \begin{equation}\label{EDMDKernel}
 \widehat{{\cal U}[\psi]}(\boldsymbol{x}) =  K(\boldsymbol{x},\bar{\boldsymbol{x}})\left( K(\bar{\boldsymbol{x}},\bar{\boldsymbol{x}}) + \sigma^2I_M \right)^{-1}  \bar{\boldsymbol{y}}
\end{equation}

It is interesting to observe that equation \eqref{EDMDKernel} is the a posteriori (Bayesian) estimate 
of the function $g(\cdot):= \psi(\boldsymbol{f}(\cdot))$, given the Gaussian Prior
$$
g(\cdot) \sim {\cal N}(0,K(\cdot,\cdot)),
$$
reconstructed from its noisy measurements $\bar{\boldsymbol{y}} = \psi( \boldsymbol{f}(\boldsymbol{x})) +  \boldsymbol{\epsilon}$.
The details of this derivation are provided in the following Section.

	\subsection{From Kernels to Koopman} \label{ssec:ker_to_koop}
	
	Let us now assume that observables $\psi(\boldsymbol{x})$ are zero mean  Gaussian processes with covariance function $K(\boldsymbol{x},\boldsymbol{x}') : = \E [\psi(\boldsymbol{x})\psi(\boldsymbol{x}')]$. This is equivalent to assume that ${\cal F}$ is a Reproducing Kernel Hilbert space with kernel $K(\boldsymbol{x},\boldsymbol{x}')$. 
	Given a generic function $\psi(\boldsymbol{x})$, its observations $\psi(\bar{\boldsymbol{x}})$ and the ``noisy'' observations
	$$	\psi(\bar{\boldsymbol{y}}) = \underbrace{\psi(\boldsymbol{f}(\bar{\boldsymbol{x}})}_{g(\bar{\boldsymbol{x}})} + \bar{\boldsymbol{\epsilon}})
$$
	we would like to estimate the function $g(\cdot)$. 
	
	Under the Bayesian framework we have
\begin{equation}\label{eq:Koopman:Bayes}
	 \hat g(\boldsymbol{x}) = K(\boldsymbol{x},\bar{\boldsymbol{x}})\left( K(\bar{\boldsymbol{x}},\bar{\boldsymbol{x}}) + \sigma^2I_M \right)^{-1}  \bar{\boldsymbol{y}}\end{equation}
	which is exactly \eqref{EDMDKernel}.
	
Let us now consider the following decomposition of the function $\psi(\boldsymbol{x})$:
\begin{equation}\label{eq:projected:psi}
\begin{array}{rcl}
\psi(\boldsymbol{x}) &=&\Pi_{\mathcal{H}_{\bar{\boldsymbol{x}}}}\left[ \psi\right](\boldsymbol{x}) + \tilde\psi(\boldsymbol{x})\\
&=& \hat\psi(\boldsymbol{x})+ \tilde\psi(\boldsymbol{x})
\end{array}
\end{equation}
where
$\Pi_{\mathcal{F}_{\bar{\boldsymbol{x}}}}\left[ \psi\right](\boldsymbol{x}) $ denotes the orthogonal projection in ${\cal F}$ of $\psi(\boldsymbol{x})$ onto the finite dimensional space ${\mathcal{F}_{\bar{\boldsymbol{x}}}}$ spanned by the kernel sections $K(:,\bar{\boldsymbol{x}})$, centered in the datapoints $\bar{\boldsymbol{x}} = ({\boldsymbol{x}}^{(1)},..,{\boldsymbol{x}}^{(M)})$.

\begin{pro} \label{pro:function}
		Given a Kernel $K(\cdot, \cdot)$ defining an RKHS $\mathcal{F}$, for any $\psi(\cdot) \in \mathcal{F}$, its projection on the space formed by the Kernel sections centred in $\bar{\boldsymbol{x}}$ is given by:
		\begin{equation*}
			\Pi_{\mathcal{F}_{\bar{\boldsymbol{x}}}} \left[ \psi(\cdot) \right] = K(\cdot, \bar{\boldsymbol{x}}) \boldsymbol{\alpha}
		\end{equation*}
		where
		$$
		\boldsymbol{\alpha}: = K(\bar{\boldsymbol{x}}, \bar{\boldsymbol{x}}) ^{-1} \psi(\boldsymbol{x})
		$$
		\end{pro}
		
		We are now ready to connect the Bayesian estimate of $g(\boldsymbol{x}) = \psi(\boldsymbol{f}(\boldsymbol{x}))$ given in \eqref{eq:Koopman:Bayes} with  the finite dimensional regularized Koopman estimator in \eqref{eq:reg:EDMD}.
		\begin{pro}
		Given an observable $\psi(\boldsymbol{x}) \in {\cal F}$, define its projection $\hat\psi(\boldsymbol{x}) : = \Pi_{\mathcal{F}_{\bar{\boldsymbol{x}}}} \left[ \psi(\cdot) \right] $. Then the regularized estimate of the Koopman operator 
		\begin{equation}\label{eq:Koopman:Kernel}
		U_f^{(r,K)} := \left( K(\bar{\boldsymbol{x}},\bar{\boldsymbol{x}}) + \sigma^2I_M \right)^{-1}  K(\bar{\boldsymbol{y}},\bar{\boldsymbol{x}})
		\end{equation}
		maps the coefficients $\boldsymbol{\alpha}$ that define    $\hat\psi(\boldsymbol{x}) = K(\bar{\boldsymbol{x}}, \bar{\boldsymbol{x}}) \boldsymbol{\alpha}$
		to 
		$$
		\boldsymbol{\beta}:= U_f^{(r,K)} \boldsymbol{\alpha}
		$$
		that define the Bayesian estimate of    $ \hat \psi( \boldsymbol{f}(\boldsymbol{x}))$ under the measurement model
		$$
		\hat\psi(\bar{\boldsymbol{y}}) = {\hat\psi(\boldsymbol{f}(\bar{\boldsymbol{x}}))} + \bar{\boldsymbol{\epsilon}}
		$$
		\end{pro}
\begin{pf}
It is sufficient to observe that  given 
$$
\hat\psi(\cdot) : =  \Pi_{\mathcal{F}_{\bar{\boldsymbol{x}}}} \left[ \psi(\cdot) \right] = K(\cdot, \bar{\boldsymbol{x}}) \boldsymbol{\alpha}
$$
we have that 
$$
\hat\psi(\bar{\boldsymbol{y}})  = K(\bar{\boldsymbol{y}}, \bar{\boldsymbol{x}}) \boldsymbol{\alpha}
$$
Hence, applying equation \eqref{eq:Koopman:Bayes} to the ``synthetic measurements'' 
$$
\hat\psi(\bar{\boldsymbol{y}}) = {{\hat\psi(\boldsymbol{f}(\bar{\boldsymbol{x}}))}}+ \bar{\boldsymbol{\epsilon}}
$$
we obtain that the Bayes estimate of ${\hat\psi(\boldsymbol{f}(\cdot))}$ is given by
$$
\begin{array}{rcl}
\widehat{\hat\psi(\boldsymbol{f}(\boldsymbol{x}))}& = &K(\boldsymbol{x},\bar{\boldsymbol{x}})\left( K(\bar{\boldsymbol{x}},\bar{\boldsymbol{x}}) + \sigma^2I_M \right)^{-1} \hat\psi(\bar{\boldsymbol{y}})\\
& = & K(\boldsymbol{x},\bar{\boldsymbol{x}})\left( K(\bar{\boldsymbol{x}},\bar{\boldsymbol{x}}) + \sigma^2I_M \right)^{-1} K(\bar{\boldsymbol{y}}, \bar{\boldsymbol{x}}) \boldsymbol{\alpha}\\
& = &  K(\boldsymbol{x},\bar{\boldsymbol{x}}) U_f^{(r,K)} \boldsymbol{\alpha}
\end{array}
$$
where we have defined 
$$
U_f^{(r,K)} : = \left( K(\bar{\boldsymbol{x}},\bar{\boldsymbol{x}}) + \sigma^2I_M \right)^{-1} K(\bar{\boldsymbol{y}}, \bar{\boldsymbol{x}})$$
The proof is concluded by observing that defining
$$
\boldsymbol{\beta}: = U_f^{(r,K)} \boldsymbol{\alpha}
$$
we have 
$$
\widehat{\hat\psi(\boldsymbol{f}(\boldsymbol{x}))} = K(\boldsymbol{x},\bar{\boldsymbol{x}})  \boldsymbol{\beta}
$$
\end{pf}

\begin{rem}
The regularized Koopman operator $U_f^{(r,K)} $ defined in \eqref{eq:Koopman:Kernel} and $U_f^{(r)} $ in \eqref{eq:reg:EDMD} simply differ in the fact that the first is expressed in the basis provided by kernel sections $K(\cdot,\bar{\boldsymbol{x}})$ while the second is written in the basis provided by the functions $\psi_i(\boldsymbol{x}) \in {\cal D}$. 
It should be observed that while in the latter case the basis functions are fixed, in the former the kernel sections depends on the observations and thus the finite dimensional approximating subspace is tuned to the observed data.
\end{rem}

	\subsection{Value-based framework}
	In the value-based perspective, given the value of an observable at the input locations $\psi(\bar{\boldsymbol{x}})$ one would like to reconstruct the values of the observable $\psi(\bar{\boldsymbol{y}})$. To this purpose the Kernel based approach  can be exploited as follows:
	\begin{enumerate}
	\item build an estimate of the observable $\psi()$ from the measurements $\psi(\bar{\boldsymbol{x}})$; this is nothing but the projection of $\psi$ on the finite dimensional space spanned by the kernel sections, i.e.
	$$
	\hat\psi(\cdot) = K(\cdot, \bar{\boldsymbol{x}}) K(\bar{\boldsymbol{x}}, \bar{\boldsymbol{x}}) ^{-1} \psi(\bar{\boldsymbol{x}})
	$$
	
	\item compute the values that this estimate takes on the mapped data $\bar{\boldsymbol{y}}$; these can be seen as ``noisy'' observations of the actual Koopman composition $\hat \psi ({\boldsymbol{f}}(\cdot))$, i.e. 
	\begin{equation}\label{eq:Koopman:value}
	\hat \psi(\boldsymbol{y}) =  K(\boldsymbol{y}, \bar{\boldsymbol{x}}) K(\bar{\boldsymbol{x}}, \bar{\boldsymbol{x}}) ^{-1} \psi(\bar{\boldsymbol{x}})
	\end{equation}
	where we can define the value-based estimate of Koopman operator on RKHS as
	\begin{equation}\label{eq:Koopman:value:kernel}
	\hat U_v^{(K)}: =  K(\boldsymbol{y}, \bar{\boldsymbol{x}}) K(\bar{\boldsymbol{x}}, \bar{\boldsymbol{x}}) ^{-1} 
		\end{equation}
 so that 
 $$
 \hat \psi(\bar{\boldsymbol{y}})=\hat U_v^{(K)} \psi(\bar{\boldsymbol{x}})
 $$
	Equation \eqref{eq:Koopman:value:kernel} can be seen as the counterpart of \eqref{DMD}, written with respect to the kernel sections.
	\item The function $ \hat \psi ({\boldsymbol{f}}(\cdot))$ can then be reconstructed from these noisy  measurements 
	following the regularised approach, i.e.
	$$
	\begin{array}{rcl}
	\widehat{\hat\psi(\boldsymbol{f}(\cdot ))} &=& K(\cdot ,\bar{\boldsymbol{x}})\left( K(\bar{\boldsymbol{x}},\bar{\boldsymbol{x}}) + \sigma^2I_M \right)^{-1} \times \\ & & \times K(\bar{\boldsymbol{y}}, \bar{\boldsymbol{x}})  K(\bar{\boldsymbol{x}}, \bar{\boldsymbol{x}}) ^{-1} \psi(\bar{\boldsymbol{x}})
	\end{array}
	$$
	\end{enumerate}
	
	Note  that the latter equation closed the gap between the value based and funtion based perspectives, providing a reconstruction of the function $\hat \psi( \boldsymbol{f}(\cdot))$ for any possible new input data.

	\section{Numerical Simulations}\label{simulations}
	In order to illustrate the effective applicability of the proposed framework, the problem of function reconstruction is actually addressed through some numerical simulations. In this section we extensively compare the estimation obtained through exploitation of kernel structure and the one based on a fixed dictionary of functions. \\
	For the sake of illustration we consider a scalar system so that transition maps and observables can be plotted and results visually inspected. We consider the  one-dimensional discrete-time autonomous state transition function as:
	\begin{equation} \label{eq:exp_sys}
		x_{t+1} = f\left( x_t \right) = - x_t + \frac{3}{\left( 1 + x_t^2 \right) } + 0.5\sin(2x_t)
	\end{equation}
	This system has an equilibrium point at $x^* \simeq 0.988$ and presents an oscillatory behaviour. \\
	We want to evaluate the performances of the different approaches both in the reconstruction of the actual state transition function $f(\cdot)$, and of a cost functional given by:
	\begin{equation*}
		J(x_t) = \Vert x_t - x_0 \Vert^2
	\end{equation*}
	This setup can be naturally embedded in the Koopman framework by considering the problem of reconstructing two different observable, given respectively by:
	\begin{equation*}
		\bar{\psi}\left( x\right) = x, \qquad \tilde{\psi}\left( x\right) = J(x)
	\end{equation*}	
	The dictionary of functions adopted in the experiments is taken as:
	\begin{equation*}
		\mathcal{D} = \left\lbrace 1, x, x^2, \sin\left( 2x\right) \right\rbrace \cup H^4
	\end{equation*}
	where $H^n$ is the set of the first $n$ Hills functions with even powers, defined as:
	\begin{equation*}
		H^n = \bigcup_{i=1}^n \left\lbrace \frac{1}{1 + x^{2i}} \right\rbrace 
	\end{equation*}
	hence $N=8$.
	It is straightforward to observe that this choice of $\mathcal{D}$ allows for a perfect reconstruction of $\bar{\psi}(\cdot)$ and so of $f(\cdot)$, while $\tilde{\psi}(\cdot)$ does not admit any representation in $\mathcal{D}$. \\
	The reconstruction restricted to the fixed dictionary is actually performed through a Kernel function $K^{(1)}(\cdot, \cdot)$, which however has been designed to yield the same computation as with the Koopman framework, in particular:
	\begin{equation*}
		K^{(1)}\left( a, b\right) = \begin{bmatrix}
			\psi_1\left( a \right) & \dots & \psi_N\left( a \right)
		\end{bmatrix} \times \begin{bmatrix}
			\psi_1\left( b \right) & \dots & \psi_N\left( b \right)
		\end{bmatrix}^\top
	\end{equation*}
	while for the true Kernel method a Gaussian RBF has been considered, defined as:
	\begin{equation*}
		K^{(2)}\left( a, b\right) = \exp\left(  -\rho \Vert a - b \Vert^2\right)
	\end{equation*}

	\begin{rem}
	In all simulation results, when computing the inverse $K(\bar{\boldsymbol{x}}, \bar{\boldsymbol{x}}) ^{-1}$, to avoid numerical problems we have added a small regularization parameter replacing $K(\bar{\boldsymbol{x}}, \bar{\boldsymbol{x}}) ^{-1}$  with $[ K(\bar{\boldsymbol{x}}, \bar{\boldsymbol{x}}) + \mu I ] ^{-1}$.
	\end{rem}
	
	In every simulation, the noise parameter $\sigma^2$ and the regularization parameter $\mu$ alluded at in the previous remark are set to:
	\begin{equation*}
		\sigma^{2(1)} = 10^{-5}, \qquad \mu^{(1)} = 10^{-5}
	\end{equation*}
	when using $K^{(1)}(\cdot, \cdot)$, only to guarantee numerical stability. \\
	On the other hand, when performing the reconstruction with $K^{(2)}(\cdot, \cdot)$, $\mu^{(2)} = 10^{-3}$ always for numerical stability, while $\sigma^{2(2)}$ is taken equal to the true variance of the actual noise injected in the state transition mapping. The $\rho$ parameter is optimized every time on a small grid. \\
	The target for the cost functional is taken as $x_0 = 0$. \\
	Data from the system are given as snapshots pair $(x_{t}, x_{t+1})$ with:
	\begin{equation*}
		x_{t+1} = f\left( x_t \right) + \omega, \quad \mathbb{V}\left[ \omega \right] = \sigma^2_T
	\end{equation*}
	for which different scenarios are taken into account, by changing the variance of the noise injection:
	\begin{equation*}
		\sigma_{T} = \left\lbrace 0, 0.2, 0.5 \right\rbrace
	\end{equation*}
	The number of available pairs is $M=50$, which consists of $5$ trajectories of length $10$. The initial point of every trajectory is sampled according to a uniform distribution between $0$ and $7$. \\
	By considering the variance of the outputs approximately equal to the variance of the inputs (which is a reasonable assumption given \eqref{eq:exp_sys}) we can understand the different settings in terms of signal-to-noise-ratio:
	\begin{equation*}
	SNR \simeq \left\lbrace \infty, 20, 8\right\rbrace 
	\end{equation*}
	\begin{figure}
		\begin{center}
			\includegraphics[width=8.4cm]{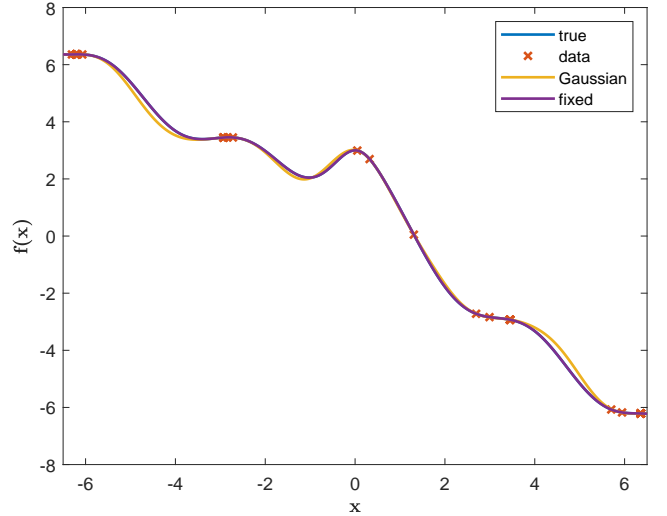}    
			\caption{Reconstruction of $\bar{\psi}(x)$ with $\sigma_{T} = 0$ and $M = 50$} 
			\label{fig:fx}
		\end{center}
	\end{figure}
	\begin{figure}
		\begin{center}
			\includegraphics[width=8.4cm]{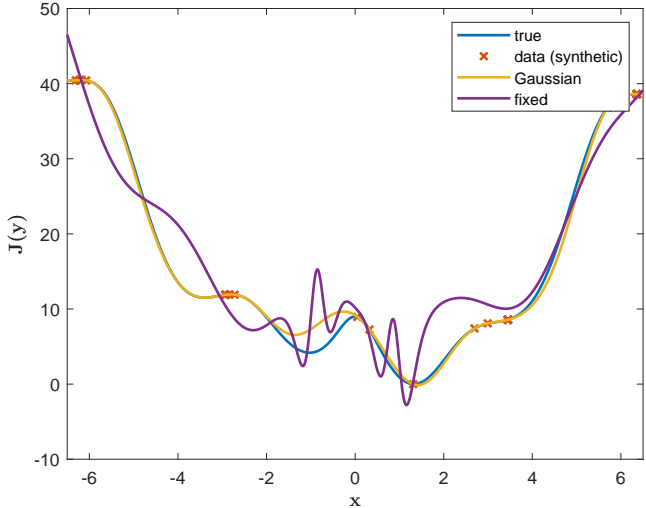}    
			\caption{Reconstruction of $\tilde{\psi}(x)$ with $\sigma_{T} = 0$ and $M = 50$} 
			\label{fig:Jy}
		\end{center}
	\end{figure}
	In figures~\ref{fig:fx}~and~\ref{fig:Jy} we can see that the two reconstructions are comparable and allow for a fair estimation of the selected observables, when neglecting the role of the noise. Clearly the identification based on the dictionary $\mathcal{D}$ will match perfectly the state transition function if no noise is injected: however the drawbacks of a reconstruction with a restricted space of functions becomes evident when dealing with a ``non-achievable'' observable, such as $\tilde{\psi}(x)$. The RBF kernel perspective is more flexible, as the estimation take place in a higher dimension, allowing for a better reconstruction. \\
	In order to better understand the behaviour of the different perspectives, we performed an extensive analysis by repeating the reconstruction with different noise realizations and evaluating every time the $L^2$ norm of the difference between the estimate and the true function, normalized by the latter, for the second observable $\tilde{\psi}(x)$.
	\begin{figure}
		\begin{center}
			\includegraphics[width=8.4cm]{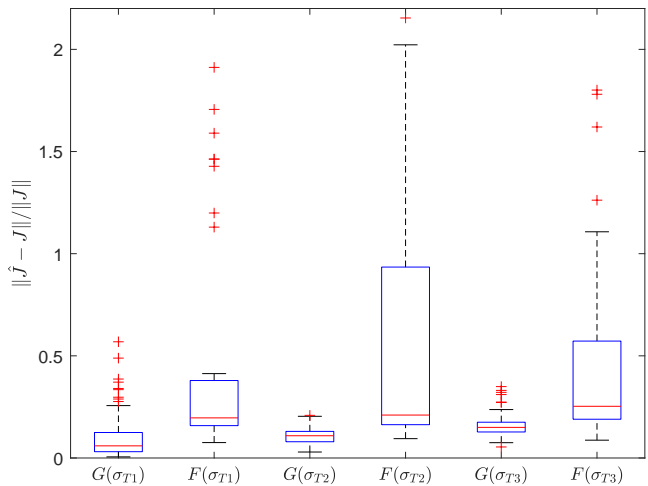}    
			\caption{Performances of the two approaches, evaluated as $\Vert \hat{J} - J\Vert/\Vert J\Vert$, for the three different noise scenarios. $G$ indicates the Gaussian kernel, $F$ the fixed basis reconstruction} 
			\label{fig:box}
		\end{center}
	\end{figure} \\
	In figure~\ref{fig:box} the results of $C = 100$ Monte Carlo simulations are presented. As already highlighted above, the Gaussian kernel yield a better overall estimation by relying on given data. The flexibility of this approach, shaping the reconstruction space according to the observed data, provides a clear advantage in the reconstruction. Clearly the performances of both perspective get worse as the SNR decreases, however the reconstruction based on the fixed dictionary is less sensitive to the variability of the noise since the behaviour of the estimated function is way more constrained.
	
	\section{Conclusions}\label{conclusions}
	In this paper we established a concrete link between Bayesian estimation with Kernel methods and the Koopman operator framework. \\
	In particular we bridged DMD and EDMD algorithms within a dual framework, that first allowed us to introduce \textit{regularization} in these finite dimensional approximation of the Koopman operator. Subsequently, by enlarging the projection space of the Koopman operator into an RKHS, we proved that the two popular data-driven procedures can be rewritten in terms of kernel sections. This naturally lead to the estimation of Koopman dynamics in RKHS, through kernel functions. \\
	This approach matches the infinite-dimensional nature of the Koopman operator but allows for a data-driven finite dimensional reconstruction. The powerful feature of the RKHS to adapt to data has been demonstrated by numerical simulations.
	
	\nocite{*}

	\bibliography{biblio}             

\end{document}